\documentclass[sigconf]{acmart}
\AtBeginDocument{%
  }

\setcopyright{none}
\acmDOI{}
\acmISBN{}
\acmConference{}{}{}
\acmYear{}
\copyrightyear{}

\usepackage{placeins}
\usepackage{float}

\begin{document}

\title{Prose2Policy (P2P): A Practical LLM Pipeline for Translating Natural-Language Access Policies into Executable Rego}


\author{Vatsal Gupta}
\affiliation{%
  \institution{Apple}
  \city{Cupertino}
  \state{California}
  \country{USA}
}
\email{vatsal_gupta@apple.com}

\author{Darshan Sreenivasamurthy}
\affiliation{%
  \institution{Apple}
  \city{Cupertino}
  \state{California}
  \country{USA}
}
\email{dtumkursreenivas@apple.com}

\begin{abstract}
Prose2Policy (P2P) is a LLM-based practical tool that translates natural-language access control policies (NLACPs) into executable Rego code (the policy language of Open Policy Agent, OPA). It provides a modular, end-to-end pipeline that performs policy detection, component extraction, schema validation, linting, compilation, automatic test generation and execution. Prose2Policy is designed to bridge the gap between human-readable access requirements and machine-enforceable policy-as-code (PaC) while emphasizing deployment reliability and auditability. We evaluated Prose2Policy on the ACRE dataset and demonstrated a 95.3\% compile rate for accepted policies, with automated testing achieving a 82.2\% positive-test pass rate and a 98.9\% negative-test pass rate. These results indicate that Prose2Policy produces syntactically robust and behaviorally consistent Rego policies suitable for Zero Trust and compliance-driven environments.
\end{abstract}



\maketitle

\section{Introduction}

Access control is foundational to security, and it has become increasingly critical with rapid digital transformation and the emergence of autonomous systems. Traditional models that hardcode authorization logic inside applications are ill-suited to today’s distributed, dynamic environments because they tightly couple decisions to code, making changes hard to update, audit, and scale, and forcing full deployment/testing cycles for every policy change \cite{adeyinka2023pac}. Consequently, organizations are shifting to externalized access control with continuous validation. As Kumar et al.\ note, Zero Trust access control frameworks rely on external policy engines and centralized verification \cite{kumar2024zerotrust}.

One of the mechanisms to externalize authorization is by implementing policy-as-code (PaC), where policies are stored as code and evaluated by a decision engine to return allow/deny outcomes for authorization requests \cite{ferraiolo2016policymachine}. Authoring these policies is a highly manual and code-centric task today, demanding expertise in languages such as eXtensible access control markup language (XACML) \cite{xacml2013}, next-generation access control (NGAC) from NIST \cite{ferraiolo2016policymachine}, Rego \cite{opa}, or Cedar \cite{cedar}. Typically, business stakeholders like analysts, auditors and managers have a better understanding of the access control requirements and can express the intent as natural language access control policies (NLACPs). However, the learning curve associated with coding these in policy languages creates a gap between human-defined intent and machine-enforced authorization. As computing systems scale across hybrid and multi-agent ecosystems, organizations require a unified framework capable of translating human intent into machine-enforceable policies.

To address this challenge, we introduce Prose2Policy, an agentic flow that interprets NLACPs, converts them into machine-enforceable Rego policies, analyzes them for quality and style (lints them), compiles them, and tests them. The agent builds on previous research by Yang et al.\ \cite{yang2025abackd} and Lawal et al.\ \cite{lawal2024llmacp} to use LLMs as the mechanism to extract policy components. It extends existing research prototypes such as RAGent \cite{jayasundara2024ragent}, ARPaCCino \cite{romeo2025arpaccino}, and AutoPAC \cite{chowdhary2025autopac} by introducing a structured, reproducible prompting framework and a feedback-enabled linting and testing mechanism to ensure real-world enforceability.

For practitioners, Prose2Policy offers a pragmatic, end-to-end path from NLACP to Rego as it extracts critical elements from NLACPs, validates them against an organization schema, and emits executable Rego with default-deny semantics plus executable unit tests. This ensures that they can pilot policy-as-code without retraining models or rewriting applications. For researchers, the tool decouples three crucial problem components in the space: policy identification (detecting which statements express access policies), component extraction (breaking down each policy into its key elements such as subject, action, resource, etc.), and executable synthesis (generating and validating policy code from these elements). This enables independent evaluation of each module and supports reproducible experiments. The pipeline’s modular prompting and validation stages provide a controlled setting to study prompting strategies, error taxonomies, and guardrail effectiveness, while the testing module enables behavior-level evaluation beyond corpus F1.

\section{Background and Related Work}

This section discusses research related to converting NLACPs into executable policies that Prose2Policy builds upon.
\subsection{Evolution and Policy Engineering}

Narouei et al.\ \cite{narouei2017topdown} noted that the manual process of translating NLACPs into formal policies is laborious, expensive, and error prone. As also highlighted by Jayasundara et al.\ \cite{jayasundara2024ragent}, natural language policy documents are inherently ambiguous, resulting in error-prone, inconsistent enforcement when manually converted to formal policy code. Similarly, Yang et al.\ \cite{yang2025abackd} demonstrated that developing policies from high-level organizational requirements is labor intensive and error prone. This led to the use of natural language processing (NLP), deep learning, and large language models (LLMs) to automate the task of converting NLACPs into machine-enforceable policies.

Early work in automated access control policy generation, such as Text2Policy proposed by Xiao et al.\ \cite{xiao2012text2policy} and ACRE from Slankas and Williams \cite{slankas2013acre}, relied on rule-based extraction using dependency structures. Later, Narouei and Takabi \cite{narouei2015roleeng} introduced top-down policy engineering frameworks leveraging semantic role labeling and neural models. Abdelgawad et al.\ \cite{abdelgawad2023abacnl} advanced this with natural-language-to-NGAC graph generation. Recent work such as AutoPAC \cite{chowdhary2025autopac} and multi-agent code-orchestrated generation \cite{khan2025multiagentiac} extends this trajectory by using LLMs to autonomously generate enforceable policies and infrastructure-as-code (IaC) configurations.

However, most previous work was not accurate enough to extract the necessary elements like subject, resource, and action from NLACPs \cite{narouei2017topdown,jayasundara2023sok,narouei2015roleeng,narouei2018ieee}. Furthermore, prior research \cite{narouei2017topdown,abdelgawad2023abacnl,yang2025abackd,lawal2024llmacp,ferraiolo2016policymachine} did not focus on extracting purpose hidden in these statements, which is a crucial component to support authorization for AI agents \cite{subramaniam2024intent}.

\subsection{LLM-Based Policy Generation}

With advances in LLMs, several studies have explored their usage to automate translation \cite{gaurav2025gaas,madan2025argen,khan2025multiagentiac,zhang2025deployability,yang2025abackd,lawal2024llmacp} with improved extraction results, but they often lack domain-specific knowledge which results in extraction of invalid subjects or resources. Some recent works \cite{jayasundara2024ragent,romeo2025arpaccino,chowdhary2025autopac} employ retrieval-augmented generation (RAG) methods that rely on organization-specific corpora to ground policy synthesis in domain context. While this improves relevance, it can reduce flexibility for large enterprises where individual application teams seek lightweight integration. Such pipelines typically need reindexing when organizational data or terminology changes, and prompt templates may require adaptation or finetuning depending on implementation.

Also, most prior research \cite{narouei2017topdown,jayasundara2023sok,narouei2015roleeng,narouei2018ieee} reported limited accuracy on component extraction and entity grounding, particularly for complex or multi-clause policies. Moreover, as Chen et al.\ \cite{chen2021codellm} and Yang et al.\ \cite{yang2025abackd} observe, LLMs may hallucinate entities, infer non-existent attributes, or yield inconsistent outputs without domain-level guardrails. These factors highlight the need for reliable, schema-aware mechanisms that ensure essential policy elements are accurately extracted and verifiable. Finally, limited research \cite{jayasundara2024ragent,mittal2025cicd} focuses on validating generated policies and improving them based on linter feedback. 

\subsection{Policy Externalization and AI Delegation}

Policy externalization has been a point of research interest with emphasis on modularity and verifiability. Xu and Zhang \cite{xu2014abacoverview} highlighted attribute-based access control (ABAC) for collaboration across multi-user and cross-organizational environments, where dynamic sharing requires fine-grained and context-aware decisions. Adeyinka \cite{adeyinka2023pac} showed that embedding authorization within applications limits scalability and auditability in hybrid systems, motivating externalized PaC frameworks. Ferraiolo et al.\ \cite{ferraiolo2016policymachine} also emphasized that separating policies is valuable in distributed systems where authorization depends on dynamic attributes and context. Governance-as-a-service \cite{gaurav2025gaas} and work by Chopra \cite{chopra2025oauth} extend this to multi-agent authorization, aligning with Prose2Policy’s conceptualization of the policy generator as a governed AI agent.

Building on this, South et al.\ \cite{south2025delegation} proposed an authenticated delegation framework extending OAuth 2.0 and OpenID Connect for AI agents, enabling verifiable delegation from humans to agents. Their work introduces the concept of translating natural-language permissions into auditable access control configurations, aligning with Prose2Policy’s objective of bridging intent and machine-enforceable authorization.

\section{Design Rationale and Technical Choices}
This section explains the key design decisions that shape our approach, specifically the prompting strategy used for translating natural-language access control requirements into executable policies and the choice of policy language.

\subsection{Prompt Engineering for Policy Extraction}

A distinguishing aspect of Prose2Policy is its structured prompting mechanism designed to address a main limitation of LLM-based policy generation: lack of control and reproducibility \cite{yang2025abackd,lawal2024llmacp,chen2021codellm}. While our modular prompting and schema/lint/test guardrails reduce hallucinations and enforce structural validity, LLM inference can remain non-deterministic: repeated runs on the same NLACP may yield different extracted representations or different (yet valid) Rego implementations. Prose2Policy therefore emphasizes deployability checks (schema validation, linting, compilation, and unit tests) and auditability of intermediate artifacts rather than guaranteeing identical outputs across runs. This prompt chaining strategy, augmented by verification feedback, encourages more consistent and interpretable outputs by constraining each step to a fixed schema and by rejecting invalid generations. Few studies have explored prompt design as a formal control mechanism. Yang et al.\ \cite{yang2025abackd} and Lawal et al.\ \cite{lawal2024llmacp} used heuristic text-to-JSON prompts, whereas Prose2Policy formalizes prompt orchestration into a schema-governed workflow with post-generation validation via Regal \cite{regal} and OPA compilation/testing. In Prose2Policy, prompt chaining supports:
\begin{itemize}
  \item \textbf{Intent identification}: specially crafted prompts based on few-shot prompting \cite{brown2020fewshot} and program-of-thought prompting \cite{chen2023pot} identify whether input represents a policy, simplify it, and extract multiple policies from provided input.
  \item \textbf{Elements extraction}: prompts extract decision, subject, action, resource, condition, and purpose (DSARCP) and return structured JSON for each recognized policy statement.
  \item \textbf{Rego synthesis}: prompts convert extracted elements into Rego policies.
\end{itemize}

\subsection{Rego as Policy Language}

Most recent research around translating NLACPs focuses on NGAC \cite{abdelgawad2023abacnl,xu2014abacoverview,ferraiolo2016policymachine,jayasundara2024ragent} and XACML \cite{narouei2017topdown,yang2025abackd,narouei2018ieee,lawal2024llmacp,xacml2013}. For our pipeline, we chose Rego due to its declarative, human-readable syntax, enabling non-technical users to validate output and provide human-in-the-loop feedback. Rego is the policy language of Open Policy Agent (OPA) \cite{opa}, which is widely used and compatible with multiple projects \cite{opaecosystem} such as Kubernetes, Envoy, Express, Terraform, and Linux-PAM. Given adoption of Rego, Prose2Policy can be useful for a large set of practitioners.

\section{System Architecture and Pipeline}

Prose2Policy is a modular and flexible tool-chain that converts natural-language access control policies into executable Rego code. The pipeline is organized into four modules:
(a) pre-processing,
(b) component extraction,
(c) schema validation, and
(d) Rego generation, refinement, and testing.
Each module can be invoked individually through the command line interface or orchestrated as a continuous flow through a user interface.

\begin{figure}[t]
  \centering
  \includegraphics[width=0.95\linewidth]{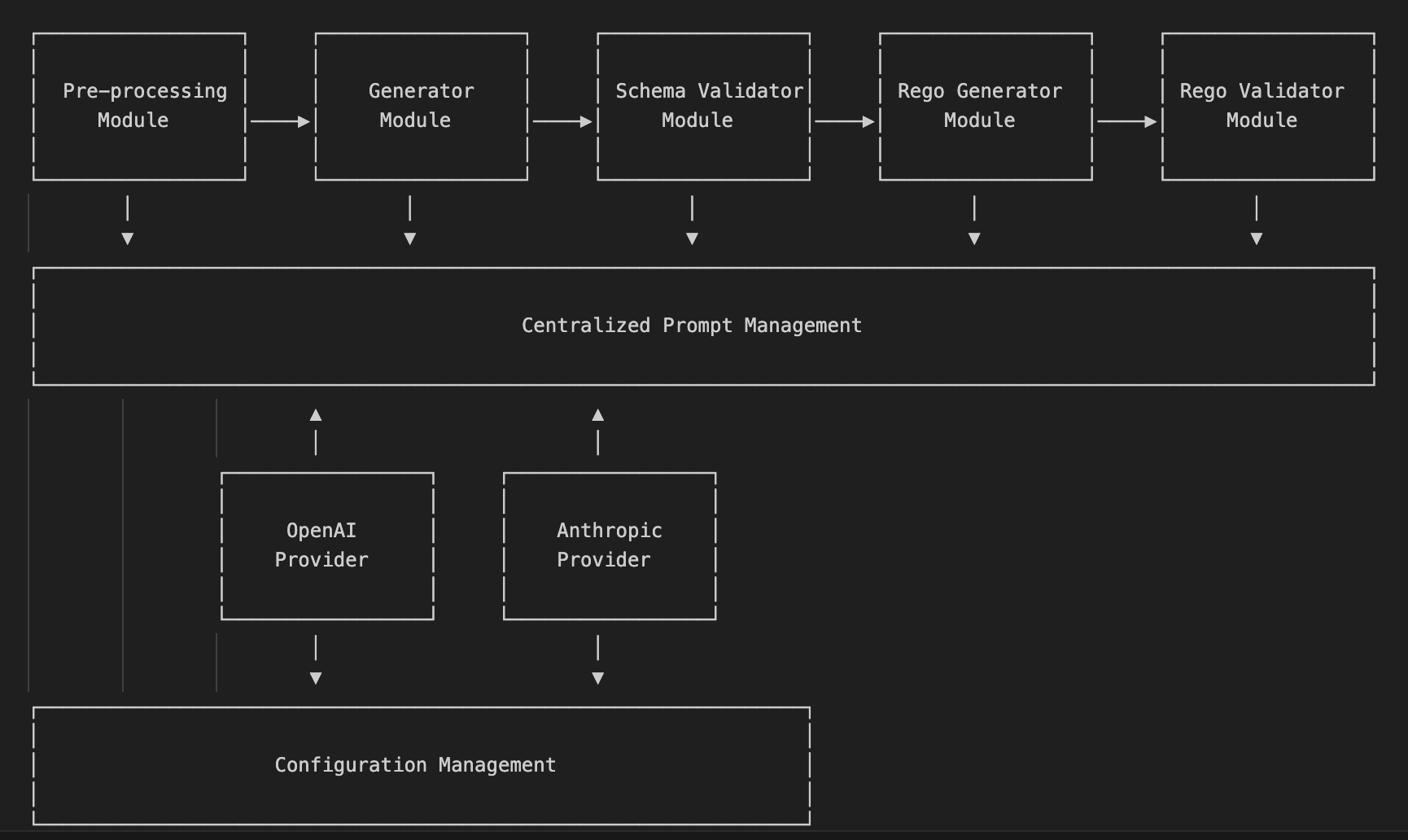}
 \Description{Architecture diagram of the Prose2Policy system showing core modules and how they are integrated}
  \caption{Prose2Policy Architecture}
  \label{fig:arch}
\end{figure}

\subsection{Pre-processing}

The pre-processing module performs:
\begin{itemize}
  \item \textbf{Policy detection} by identifying if the statement represents one or multiple policies. The detection logic uses an explicit definition: an access control policy describes \emph{who} (subject) can or cannot perform an \emph{action} on a \emph{resource}, sometimes under \emph{conditions} or for a \emph{purpose}. This filtering improves precision during pre-processing.
  \item \textbf{Co-reference resolution} by resolving pronouns and implicit references.
  \item \textbf{Text segmentation} by breaking input text into multiple policy statements.
\end{itemize}

By combining co-reference resolution and text segmentation, an input such as ``Nurses are allowed to read prescriptions, but they are not allowed to change them'' becomes two NLACPs: ``Nurses are allowed to read prescriptions'' and ``Nurses are not allowed to change prescriptions.''

Unlike Text2Policy \cite{xiao2012text2policy} and ACRE \cite{slankas2013acre}, which rely on handcrafted syntactic patterns, Prose2Policy couples prompt-based intent detection with deterministic normalization, providing a transparent pre-processing audit trail.

\subsection{Component Extraction}

The extraction module identifies core access control components using prompts to an LLM following few-shot mechanisms. The system extracts:
\begin{itemize}
  \item \textbf{Decision}: allow/deny determination (e.g., "allow", "permit", "deny")
  \item \textbf{Subject}: actor to whom the policy applies (e.g., "administrators", "users")
  \item \textbf{Action}: operation being regulated (e.g., "access", "modify", "view")
  \item \textbf{Resource}: protected asset (e.g., "database", "files", "records")
  \item \textbf{Condition}: contextual constraints e.g., "during business hours", "with approval")
  \item \textbf{Purpose}: intended goal of the action(e.g., "for maintenance", "for auditing")
\end{itemize}

\subsection{Schema Validation}

The schema validation module ensures extracted DSARCP components conform to predefined organization-specific schemas (JSON) listing valid values for each component. If crucial components like subject, action, or resource are not identified, the flow stops and missing attribute details are shared with the policy administrator. This stage is configurable and can be disabled for research use. Prose2Policy includes multiple schemas based on datasets used in prior research \cite{jayasundara2024ragent} which can be modified as needed.

\subsection{Rego Generation, Refinement, and Testing}

The generator emits Rego modules that follow a deny-by-default pattern and annotates rules with extracted DSARCP for auditability. If multiple policy statements are identified, the output is a single Rego module with all statements encoded.

Generated Rego is validated by a Rego linter and style checker (Regal) \cite{regal}; feedback is incorporated to improve generated policy code. Policies are then compiled to ensure OPA compatibility and tested against dynamically generated positive and negative test cases executed via \texttt{opa test}. Unit tests are automatically generated by the system based on extracted policy components, minimizing manual intervention. This testing approach provides confidence in generated policies before deployment. While RAGent \cite{jayasundara2024ragent} and ARPaCCino \cite{romeo2025arpaccino} emphasize generation quality and policy feedback, Prose2Policy focuses on deployment reliability by integrating static checks and unit tests.

\section{Prose2Policy Features}

Prose2Policy provides a web-based interface that unifies the modular pipeline into an interactive environment. The design supports practitioners who author, validate, and deploy policies, and researchers who experiment with prompting, evaluation, and schema configuration.

\subsection{Understanding and Experimenting with the Flow}

The single-policy view serves as an exploratory environment for understanding how natural language statements evolve into executable Rego policies. A user can paste a statement and observe each stage from policy detection and DSARCP extraction to schema validation and code synthesis. The interface displays intermediate reasoning and explanations for transparency.

Practitioners can use this mode to fine-tune access rules and verify logic, while researchers can analyze prompt behavior and evaluate extraction accuracy. The same interface allows editing outputs generated by modules, enabling controlled experimentation with alternative phrasing, few-shot examples, or context windows.

\subsection{Batch Processing and Corpus Preparation}

The batch processing feature enables users to run Prose2Policy across large text or groups of policy statements, following the configured flow for all statements. This is valuable for teams onboarding to policy-based access control or extracting policies from design docs, user stories, or functional requirement documents, as highlighted by Lawal et al.\ \cite{lawal2024llmacp}. Users can download generated Rego modules and import them into version control systems. For researchers, this tab supports experiments comparing prompting variants or model configurations across identical datasets.

\subsection{Testing and Validation of Generated Policies}

Once policies are generated, the testing and validation feature verifies correctness and deployment readiness. Prose2Policy runs each policy through \texttt{opa check} and linting (Regal) followed by unit tests executed using \texttt{opa test}. It generates positive and negative test cases derived from DSARCP components. This test suite bridges generation accuracy with runtime semantics, ensuring policies are both syntactically valid and operationally sound.

Prose2Policy offers users the choice between LLM-based and Rule-based test generation modes. While both modes are effective for standard DSARCP structures, test generation faces challenges with policies referencing external constants, nested or compound conditions, contextual variables, type conversions, or time-based evaluations. These limitations impact test generation logic rather than the correctness of generated Rego code. The LLM-based mode provides improved handling of some complex cases, but certain challenges persist in both modes.

\subsection{Configuration and Prompt Customization}

The configuration tab provides control over how policies are extracted, validated, generated and tested. Users can select among LLM providers based on cost, accuracy, or privacy requirements,  can toggle schema validation and Rego validation, and can also decide if they want to use LLM or rules to generate test cases. Users can also customize prompts by editing prompt templates to refine behavior for specific domains or datasets. This makes Prose2Policy adaptable for production and extensible for research experimentation without modifying the core engine.

\section{Evaluation and Results}

Prose2Policy was evaluated on the ACRE dataset \cite{slankas2013acre}, which has been used in prior research such as Narouei et al.\ \cite{narouei2017topdown} and RAGent \cite{jayasundara2024ragent}. We used the output file generated by RAGent, consisting of 485 access control statements, as input for our tool.

Because Prose2Policy filters out ambiguous statements without explicit authorization intent, it accepted 389 statements as valid NLACPs and generated Rego policies for them. It is worth noting that the RAGent output file contains only access control statements, and our pipeline applies stricter filtering to identify valid NLACPs, resulting in a lower acceptance rate but higher policy realism. Furthermore, the output formats of prior tools such as RAGent differ from Prose2Policy: previous work typically focused on policy identification or extraction and did not generate Rego code or executable unit tests. As a result, a direct, line-by-line comparison of generated Rego policies or test results is not possible. Of these, 18 policies failed to produce valid Rego due to syntax-related issues introduced during automated generation, resulting in 371 successfully compiled policies leading to 95.3\% (371/389) compile rate.

Each policy was evaluated using automatically generated positive and negative unit tests derived from its DSARCP elements. With the LLM-driven test generation mechanism, Prose2Policy achieved a 82.2\% positive-test pass rate (305/371) and a 98.9\% negative-test pass rate (367/371). When compared with rule-based test cases, Prose2Policy achieve 62.1\% positive-test pass rate and a 97.1\% negative-test pass rate. The high negative-test pass rate demonstrates strong enforcement of deny-by-default semantics, while the positive-test failures primarily reflect limitations of automated test generation for more complex policies, rather than defects in the generated Rego code itself.

\section{Conclusion and Future Work}

Prose2Policy democratizes the translation of natural-language access requirements into executable, auditable policies. Its modular pipeline enables both precision and adaptability in enterprise environments while allowing researchers to improve the performance and accuracy of the translation process. In uniting these two audiences, Prose2Policy functions as both a policy-as-code workbench and a research testbed. It lays a foundation for reproducible experiments and practical deployment.

Future enhancements will focus on improving adaptability, scalability, and reliability:
\begin{itemize}
\item \textbf{Advanced schema support}: We plan to enhance schema validation to support wildcards, pattern-based matching, and hierarchical structures, making the tool more scalable for large or dynamic domains and reducing the need for manual enumeration of values.
\item \textbf{Independent test generation}: To further increase trust in automated validation, we aim to decouple policy and test generation, potentially by using separate LLMs or human-in-the-loop processes, to reduce the risk of correlated errors.
\item \textbf{Enhanced support for complex policies}: We will improve the robustness of test generation and policy synthesis for policies with nested conditions, contextual variables, or references to external constants.
\item \textbf{Multi-language output}: We plan to extend support beyond Rego to other policy languages such as Cedar, XACML, and NGAC, enabling broader adoption across diverse access control ecosystems.
\item \textbf{Deterministic and semantically stable outputs}: We plan to reduce run-to-run variability by enforcing canonical Rego templates (e.g., input-injected context such as time) and by evaluating semantic equivalence across repeated runs.
\item \textbf{Feedback-driven refinement}: Finally, we aim to implement a feedback loop that uses test outcomes and deployment feedback to automatically refine subsequent policy generations.
\end{itemize}

\clearpage
\FloatBarrier
\bibliographystyle{ACM-Reference-Format}
\bibliography{main}

\appendix
\FloatBarrier
\section{Additional Resources and Details}
This appendix provides supplementary material intended to support reproducibility and does not introduce new technical contributions.

\subsection{Example NLACP to Rego Translation}

The diagram shows how a policy statement goes through the different modules, along with the inputs and outputs for each module.

\begin{figure}[H]
  \centering
  \includegraphics[width=0.95\linewidth]{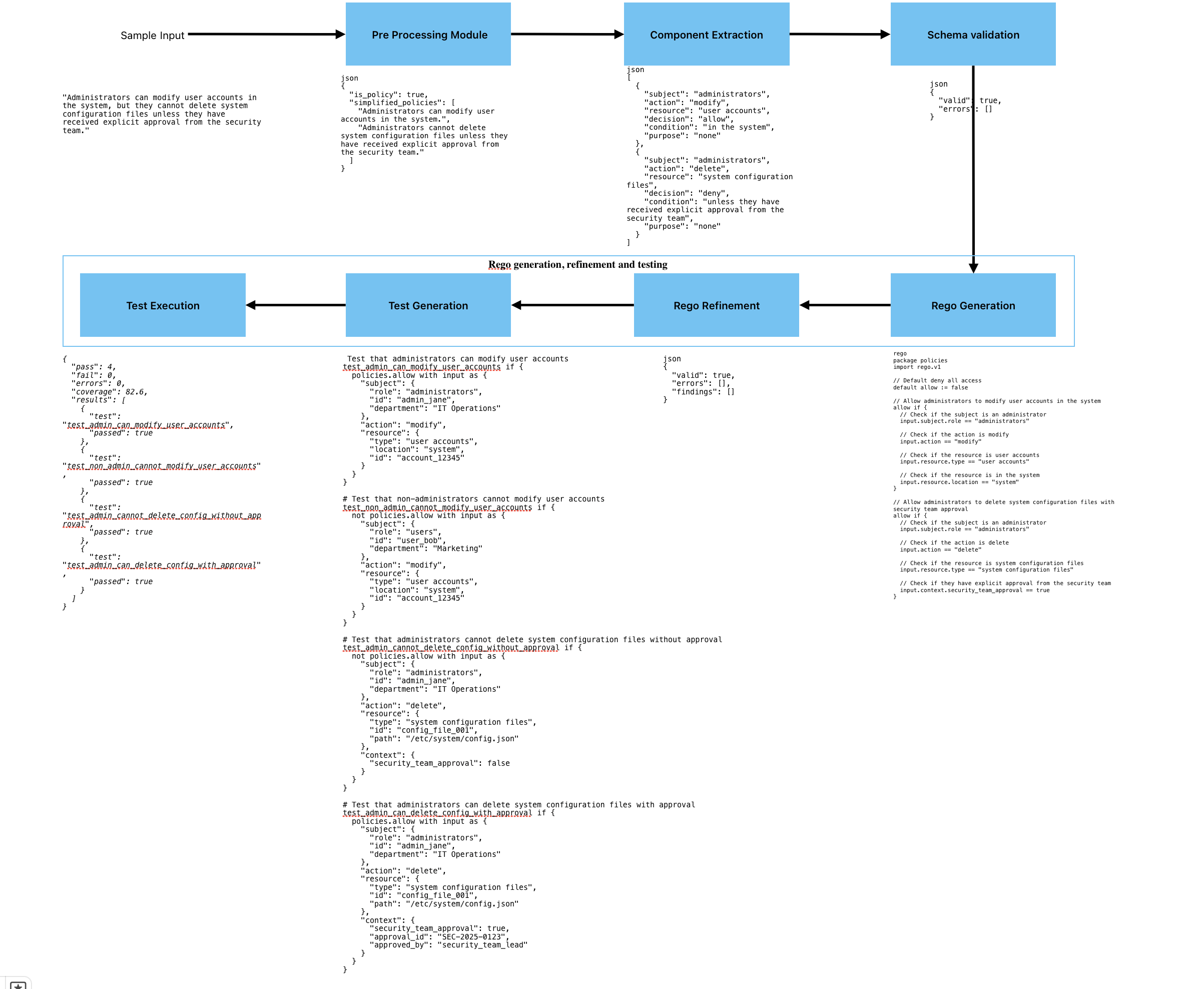}
  \Description{Diagram illustrating how a natural-language policy flows through Prose2Policy modules with corresponding inputs and outputs.}
  \caption{Inputs and outputs of each component}
  \label{fig:flow}
\end{figure}

\textbf{Note}: Code for the tool described in this paper will be made publicly available upon acceptance, in accordance with SACMAT policy.

\end{document}